\documentclass[letterpaper, 10 pt, conference]{ieeeconf}

\IEEEoverridecommandlockouts
\usepackage{graphicx}
\usepackage{graphics} 
\usepackage{epsfig} 
\usepackage{times} 
\usepackage{amsmath} 
\usepackage{amssymb}  
\usepackage[table]{xcolor}
\usepackage[ruled,linesnumbered, noend]{algorithm2e}
\usepackage{flushend} 

\usepackage{tabulary}
\newcolumntype{K}[1]{>{\centering\arraybackslash}p{#1}}

\usepackage{soul}
\usepackage{cite}
\usepackage{comment}
\usepackage{multirow}
\usepackage{graphicx}
\usepackage{wrapfig}
\usepackage{soul, color}
\usepackage{wrapfig}
\usepackage{mathtools}
\usepackage{graphicx}
\usepackage{booktabs}
\usepackage{adjustbox}

\usepackage{hyperref}
\hypersetup{
    colorlinks=true,
    linkcolor=black,
    filecolor=magenta,      
    urlcolor=blue,
    pdftitle={Overleaf Example},
    pdfpagemode=FullScreen,
    }

\usepackage{caption}
\usepackage{subfigure}

\usepackage{lipsum}  

\newcommand{\ve}[1]{\mathbf{#1}} 

\title{IMU-based Real-Time Crutch Gait Phase and Step Detections in Lower-Limb Exoskeletons}
\author{Anis R. Shakkour, David Hexner, Yehuda Bitton and Avishai Sintov
\thanks{A. R. Shakkour and A. Sintov are with the School of Mechanical Engineering, Tel-Aviv University, Israel. E-mail: anisshakkour@mail.tau.ac.il; sintov1@tauex.tau.ac.il.}
\thanks{D. Hexner is with LifeWard Ltd., Yokneam Ilit, Israel; Y. Bitton is with Binata Ltd. Yokneam Ilit, Israel. }
\thanks{This research was supported by the Israel Innovation Authority (Grant No. 77857).}
}

\begin{document}

\setlength{\belowdisplayskip}{2pt}
\setlength{\belowdisplayshortskip}{3pt}
\setlength{\abovedisplayskip}{2pt} 
\setlength{\abovedisplayshortskip}{3pt}
\setlength{\parskip}{0pt}


\maketitle
\thispagestyle{empty}
\pagestyle{empty}



\begin{abstract}
    Lower limb exoskeletons and prostheses require precise, real time gait phase and step detections to ensure synchronized motion and user safety. Conventional methods often rely on complex force sensing hardware that introduces control latency. This paper presents a minimalist framework utilizing a single, low cost Inertial-Measurement Unit (IMU) integrated into the crutch hand grip, eliminating the need for mechanical modifications. We propose a five phase classification system, including standard gait phases and a non locomotor auxiliary state, to prevent undesired motion. Three deep learning architectures were benchmarked on both a PC and an embedded system. To improve performance under data constrained conditions, models were augmented with a Finite State Machine (FSM) to enforce biomechanical consistency. The Temporal Convolutional Network (TCN) emerged as the superior architecture, yielding the highest success rates and lowest latency. Notably, the model generalized to a paralyzed user despite being trained exclusively on healthy participants. Achieving a 94\% success rate in detecting crutch steps, this system provides a high performance, cost effective solution for real time exoskeleton control.
\end{abstract}

\section{Introduction}

Lower-limb exoskeletons are transformative robotic devices that either assist in rehabilitation \cite{Foroutannia2026} or restore standing and walking ability to individuals with severe mobility impairments, such as those resulting from spinal cord injury \cite{Baud2021}. 
Similarly, active lower-limb prostheses enable amputees to restore mobility and achieve a more natural gait \cite{Asif2021}. These assistive devices are fundamentally designed to enhance user safety, promote smooth and natural motion, and maximize comfort. However, optimal performance is unattainable without a robust and functional gait phase detection method \cite{Vu2020}. This detection unit is essential, as it provides the device's main controller with real-time information about the user's current gait phase. By accurately detecting the current phase (e.g., heel strike, swing), the device can produce the corresponding, temporally synchronized motion and actuation functionality. Consequently, integrating a highly accurate and low-latency gait detection strategy is paramount for improving the overall performance, responsiveness, and safety of the control system in the walking device.

\begin{figure}
    \centering
    \includegraphics[width=0.75\linewidth]{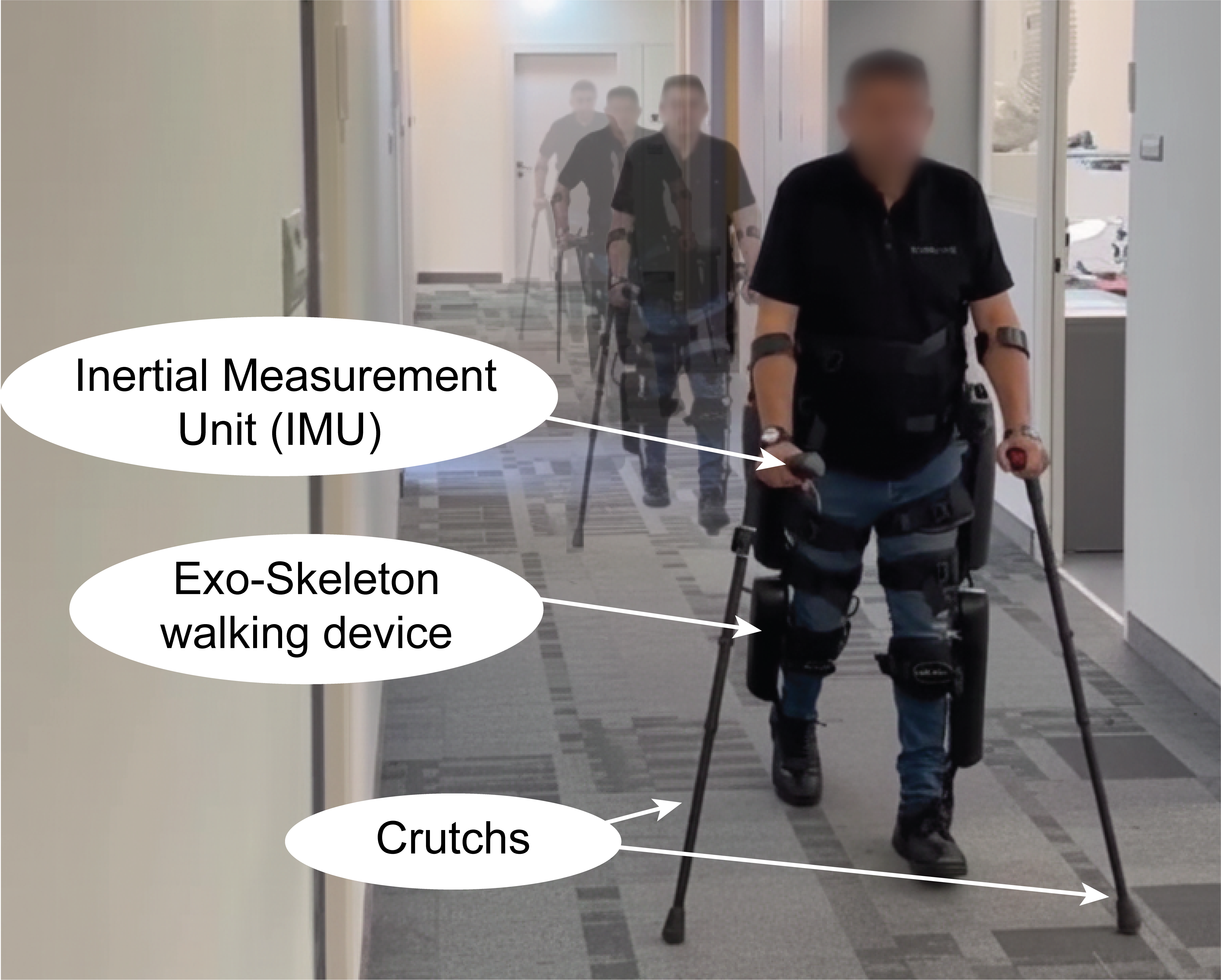}
    \vspace{-0.2cm}
    \caption{\small A participant walking with the ReWalk \cite{Esquenazi2012} lower limb exoskeleton and crutches. The right-hand side crutch is equipped with an Inertial Measurement Unit (IMU) for gait phase detection.}
    \label{fig:intro}
    \vspace{-0.6cm}
\end{figure}

Current methods for gait phase detection typically employ various sensor modalities. The detection strategies can be clearly separated based on the location of the sensing hardware. Sensors mounted directly on the robotic device commonly include joint encoders and torque sensors to measure the robot's state, and force sensors integrated into the foot sole to detect ground contact \cite{Caldas2017, Park2020}. While these provide direct kinematic and kinetic feedback, they primarily measure the current device state and do not interact with the user to capture intent \cite{Baud2021}. This reliance on hindsight data precludes intent recognition, as the system only observes motion after sensing a shift in the body center of mass. Such reactive sensing fails to detect intent, resulting in sluggish coordination that lacks the proactive responsiveness required for natural walking. Furthermore, relying solely on foot sensors can lead to inconsistent performance across varied terrains or for users with atypical gait patterns. A comprehensive, yet challenging, approach involves Electro-Myography (EMG), which detects muscle activation to predict intent, but its real-world utility is limited by electrode placement sensitivity, signal noise, and inter-subject variability \cite{Kim2022, Guerra2024}. 

An alternative and often more proactive approach utilizes sensors placed on the assistive devices, primarily the crutches, which the user manipulates to express walking intent \cite{Lancini2016, Li2021}. In addition to assisting in balance, the crutch acts as the primary mechanical interface between the user's intention and the ground. Instrumented crutches typically incorporate force sensors to directly quantify the axial load and shear forces exerted by the user onto the crutch, specifically measuring the interaction forces between the crutch tip and the ground \cite{Chamorro2016, Fong2022}. An advanced and highly functional approach integrates these kinetic measurements with Inertial Measurement Units (IMUs) \cite{Narváez2022, Narvaez2024}. This sensor fusion provides a comprehensive dataset for accurately modeling the crutch's state, providing a crucial predictive window for the exoskeleton controller \cite{Jung2012}. Crucially, while IMUs are easy to deploy, the viability of relying solely on the crutch's inertial data, without any supporting force sensors, for robust, multi-phase gait classification has not yet been thoroughly explored, presenting a significant opportunity for simplification and cost reduction.

This gap motivates our work. In this paper, we explore the use of an IMU on a crutch for cost-effective gait phase and step detections. We present a novel, minimalist, and computationally efficient framework for crutch gait detection tailored for real-time exoskeleton control (Figure \ref{fig:intro}). Our key contribution lies in demonstrating that gait detection can be achieved using signals from a single, low-cost and standard IMU concealed within on the crutch hand grip, completely eliminating the need for force sensors or any mechanical modification to the crutch tip. While the common approach for control systems is to classify the standard four phases of the crutch gait cycle (i.e., stance, take-off, swing and strike \cite{Costamagna2017}), a more robust controller requires the addition of a crucial functional phase. We extend the model to a five-phase system by adding the \textit{Auxiliary} phase covering any non-gait activities such as standing, drinking, scratching or opening a door. This inclusion is vital for safety, as this phase allows the control system to enter a locked state, preventing undesired motion caused by incidental sensor readings during non-locomotor activities. We consider a temporal-based classification model to achieve accurate and low-latency performance. The resulting methodology offers a high-performance and cost-effective solution to significantly improve human-exoskeleton coordination and responsiveness.



\section{Method}

\subsection{System}

\begin{figure}
    \centering
    \includegraphics[width=0.5\linewidth]{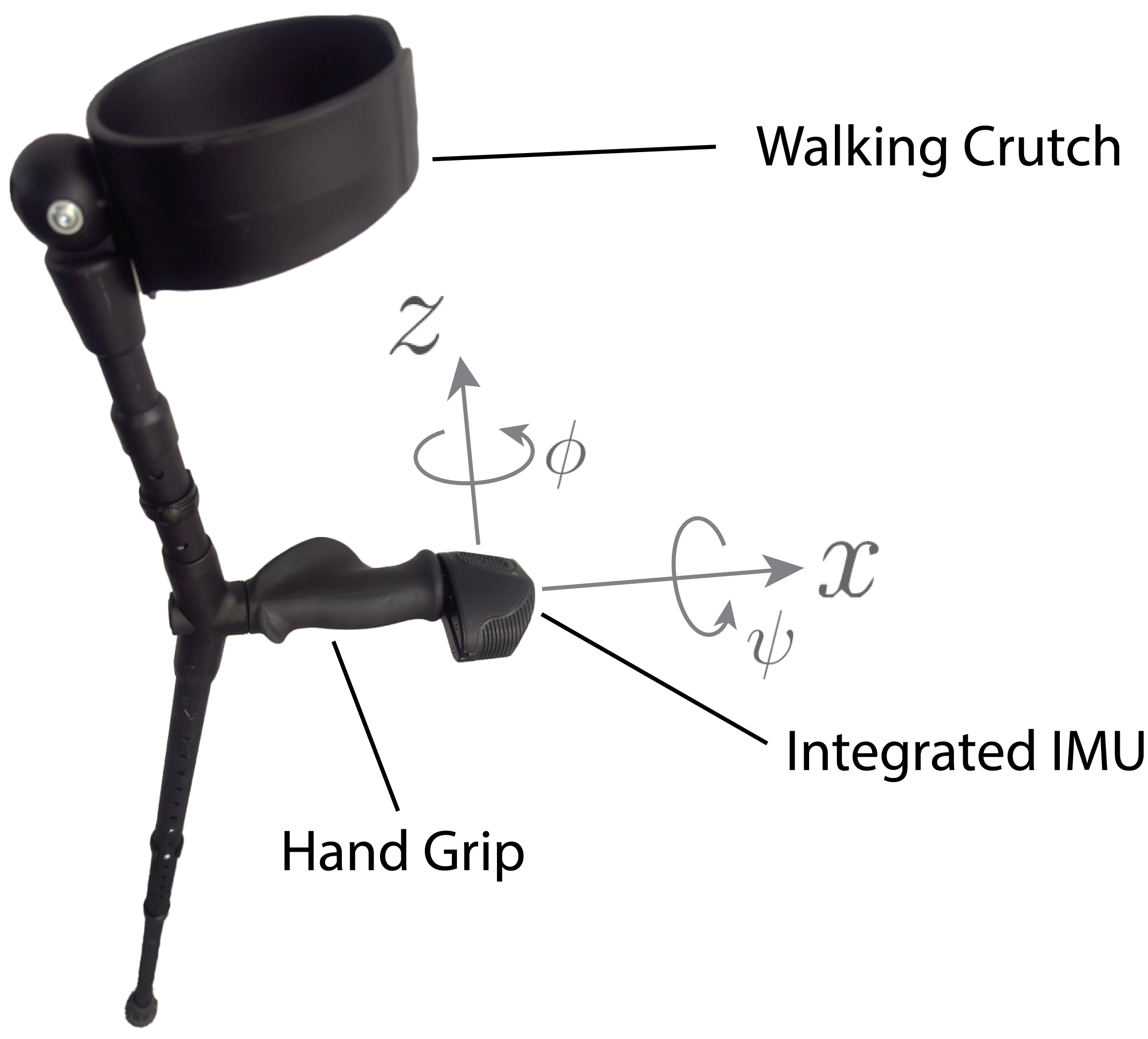}
    \caption{\small The crutch with an integrated IMU within the hand grip.}
    \label{fig:crutch}
    \vspace{-0.5cm}
\end{figure}

In this work, the crutch system is the dedicated interface for the detection of user gait intent, utilizing a standard forearm crutch instrumented with a single, low-cost off-the-shelf IMU. The IMU was integrated within the hand grip of the right hand side crutch as seen in Figure \ref{fig:crutch}. Specifically, the $z$-axis is oriented vertically upward, aligned with gravity during standing, the $x$-axis is pointed horizontally toward the walking direction and the $y$-axis is pointed perpendicular to the user's walking direction. This specific alignment ensures that the output signals, such as forward pitch rotation about the $x$-axis, directly and naturally correspond to key crutch motions like the swing phase.

The IMU sensor provides streaming at 100 Hz including all nine degrees of freedom: tri-axial linear acceleration ($a_x$, $a_y$, $a_z$), tri-axial angular velocity ($\omega_x$, $\omega_y$, $\omega_z$) and tri-axial magnetic field strength ($\mu_x$, $\mu_y$, $\mu_z$). Additionally, the IMU's internal sensor fusion engine provides pre-processed orientation outputs in both Euler angles and quaternions, offering flexibility in choosing input features for the classification model. Consequently, a processed measurement vector  at time $t$ is defined by $\ve{x}_t \in \mathbb{R}^9$. This vector is composed of the tri-axial raw linear acceleration, filtered angular velocity and fused orientation angles, represented as
\begin{equation}
    \ve{x}_t = (\tilde{a}_x, \tilde{a}_y, \tilde{a}_z, \tilde{\omega}_x, \tilde{\omega}_y, \tilde{\omega}_z, \psi, \theta, \phi)^T
    \label{eq:state}
\end{equation}
where $\tilde{a}_x$, $\tilde{a}_y$ and $\tilde{a}_z$ are the unbiased accelerations acquired by rotating $a_x$, $a_y$ and $a_z$ into the global coordinate system using the orientation quaternions. This transformation allows separation of gravity and motion components, producing the global unbiased accelerations. Similarly, velocity $\tilde{\omega}_j$ denotes the angular velocity components ($\omega_j$) after being smoothed by a low-pass filter to ensure a clean and robust signal for the model. Parameters $\psi$, $\theta$, and $\phi$ represent the Euler angles derived from the IMU’s internal sensor fusion engine, which integrates gyroscope and magnetometer data. Utilizing these processed Euler angles instead of raw magnetometer signals significantly improves system reliability by mitigating the impact of local electromagnetic interference and magnetic anomalies.

The entire system is made portable by a rechargeable Li-Po battery, offering sufficient power for continuous operation for more than 6 hours on a single charge, which is adequate for typical testing or extended daily use sessions. Raw data stream from the IMU is wirelessly transmitted to a nearby computer over Bluetooth Low Energy (BLE), enabling easy data collection. In real-time operation, the data can be transmitted directly to the embedded exoskeleton controller for analysis and inference. 


\subsection{Problem Definition}

The objective of this research is to formulate a real-time classifier of gait phases which imply about intent during crutch-assisted ambulation, specifically targeting lower-limb exoskeleton control. Given some IMU measurement $\ve{I}_t$ acquired at time $t$, we aim to classify the user's current gait status into one of $m=5$ discrete classes, $\{\mathcal{C}_1, \ldots, \mathcal{C}_5\}$. These five functional classes collectively represent the full crutch-assisted gait cycle and all necessary non-locomotor states. The phases are defined as: $\mathcal{C}_1$: \textit{Stance}, where the crutch is in full contact with the ground, bearing the user's weight and maintaining stability; $\mathcal{C}_2$: \textit{Take-off}, which is the instant the crutch tip lifts off the ground, marking the end of the stance phase and preparation for movement; $\mathcal{C}_3$: \textit{Swing}, where the crutch is off the ground and moving forward toward the next contact point; $\mathcal{C}_4$: \textit{Strike}, the instant the crutch tip makes contact with the ground to initiate the load-bearing stance phase; $\mathcal{C}_5$: \textit{Auxiliary}, representing all non-locomotor states where the user has halted forward motion. This encompasses both active standing, where the user maintains a stable, stationary posture with grounded crutches, and the performance of static tasks unrelated to walking, such as drinking or hand washing. The first four phases are within the standard crutch gait cycle, and their detection enables the synchronization of the walking device based on the user's upper-body intent. Phases $\mathcal{C}_5$ is essential to signal the control system to behave accordingly and not initiate undesired gait steps due to incidental crutch movements during non-locomotor activities.

Beyond instantaneous phase classification, the core problem involves detecting entire completed steps and their corresponding temporal lengths. A step is identified through the successful sequential transition from $\mathcal{C}_1$ to $\mathcal{C}_4$. By accurately predicting the start and end of these cycles, we can evaluate the system's temporal precision, ensuring the exoskeleton mirrors the user's actual step duration. Therefore, we aim to design and train a classifier $\mathcal{F}(\mathbf{I}_t)$ that accurately outputs the current gait class $\mathcal{C}_i$ in real-time. The system must effectively distinguish between the similar kinematic patterns of the locomotion phases ($\mathcal{C}_1$–$\mathcal{C}_4$), identify critical non-locomotor states ($\mathcal{C}_5$), and reliably segment entire steps to ensure the high-level control of the exoskeleton is both responsive and temporally accurate.


\subsection{Data Collection}

A dedicated dataset $\mathcal{H}$ was collected for training and evaluating the gait phase recognition models. Several human subjects were recruited from two distinct groups to ensure the model's generalizability: Healthy able-bodied subjects and an exoskeleton user who rely on the ReWalk device for mobility. To standardize the recordings, all subjects followed a 15 m path, turned, and walked back to the starting point, repeating this cycle until the session was complete. This design ensured each trial introduced natural variability into the dataset, encompassing straight walking, turning, and stopping phases, and incorporating diversity in crutch-assisted locomotion patterns. The subjects were also asked to arbitrary induce the auxiliary task ($\mathcal{C}_5$) phase during the sessions. While no specific instructions were given to the subjects on how to walk, subjects employed multiple common crutch gait strategies, including Two-Point Gait (where the crutch swing preceded the contralateral leg step), Swing-To Gait (where both crutches were swung forward together, followed by advancing both legs to the level of the crutches), and Swing-Through Gait (where both crutches were swung forward, followed by moving both legs beyond the crutch tips). To maintain a robust and practical training resource, an effort was made to achieve a near-uniform distribution of samples across all five classes.

Finally, the collected data was pre-processed to include labeled sequences. Sessions were truncated to multiple sequences where each sequence $\ve{I}_t$ is represented as a sliding window of length $h$. This window aggregates the measurement vectors \eqref{eq:state} over a specific temporal context, providing the model with the necessary dynamics to distinguish between gait phases. The sliding window at time $t$ is defined as a tensor:
\begin{equation}
    \ve{I}_t=[\ve{x}_{t-h+1},\ve{x}_{t-h+2},\ldots,\ve{x}_t].
\end{equation}
Tensor $\ve{I}_t \in \mathbb{R}^{h \times 9}$ serves as the input for each inference step. This windowing approach allows the models to evaluate not just a static posture, but the kinematic trajectory of the crutch, which is essential for identifying transitional events like Strike or Take-off. Sequence tensor $\ve{I}_t$ was manually labeled with the class $c_i\in\{1,\ldots,5\}$. The labeling was performed off-line by an expert observing the cyclic signals and synchronized session videos. Consequently, the resulting training dataset is composed of synchronized, labeled time-series sequences of the form $\mathcal{H}=\{(\ve{I}_i, c_i)\}_{i=1}^N$, where $\ve{I}_i$ is the IMU data sequence $i$ and $c_i$ is the corresponding gait phase label. 


\subsection{Gait Phase Detection Model}

The final stage of our methodology involves the sequence classification model $\mathcal{F}$, which is tasked with mapping a given IMU time-series feature window $\mathbf{I}_t$ to one of the five required gait phase classes in real-time. Since the system is designed for embedded real-time control where processing overhead and power consumption must be minimized, we prioritized lightweight, resource-efficient models. To identify the model that provides the optimal balance between classification accuracy and minimal inference latency, we performed a comparative benchmark across three low-complexity models: the Long Short-Term Memory (LSTM) network \cite{Yu2019}, the Temporal Convolutional Network (TCN) \cite{Mizrahi2024}, and the Transformer \cite{Vaswani2017}.  

The LSTM network is evaluated as the baseline recurrent architecture, inherently designed to capture long-term temporal dependencies within the IMU sequence window. The TCN is selected for its demonstrated efficacy in time-series analysis, employing causal and dilated convolutions to efficiently model long-range temporal context with a large receptive field while benefiting from the parallelization capabilities of convolutions, making it computationally efficient. Finally, the Transformer network, leveraging its powerful self-attention mechanism, is included to assess its ability to identify global dependencies and weigh the importance of different time-steps across the entire input window. Each model was implemented with a consistent input structure receiving the feature tensor, followed by its respective sequence processing core, and culminating in a final dense layer with softmax activation.

\begin{figure}
    \centering
    \includegraphics[width=0.55\linewidth]{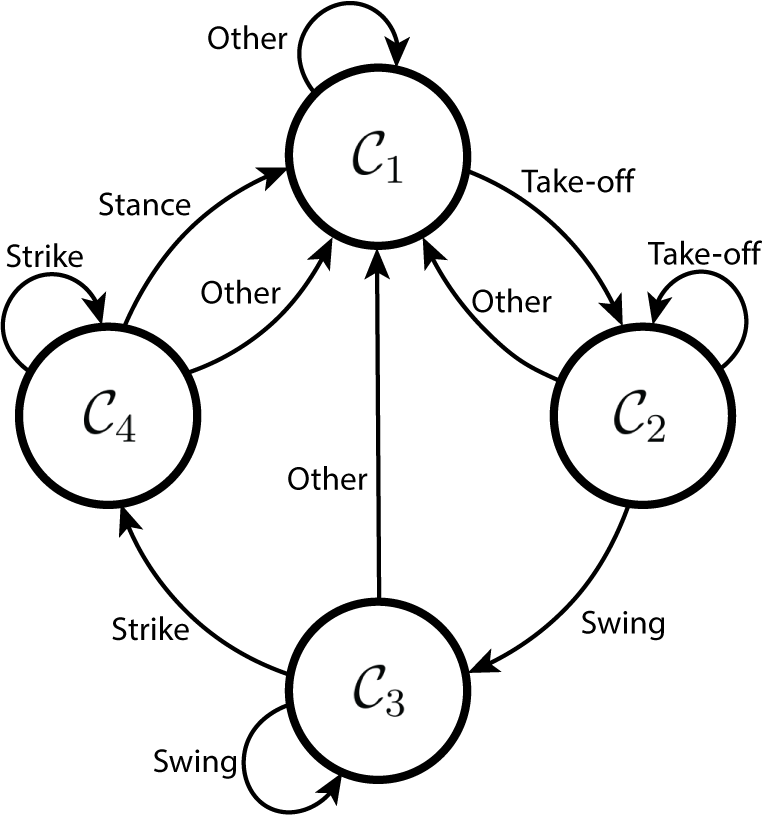}
    \vspace{-0.2cm}
    \caption{\small Finite State Machine (FSM) for crutch gait phase refinement and step event detection.}
    \label{fig:FSM}
    \vspace{-0.6cm}
\end{figure}

The softmax activation generates a probability distribution across the five defined gait classes. To refine these raw predictions into a coherent gait trajectory, the decoded phase sequence is integrated with a Finite State Machine (FSM) designed to identify individual crutch steps and enforce biomechanical consistency. This FSM acts as a causal supervisor that monitors for the progression of a single step, specifically tracking the Take-off followed by a subsequent plausible sequence of gait phases. The FSM scheme for the crutch gait cycle is illustrated in Figure \ref{fig:FSM}. 
By operating casually in time, the FSM can be applied identically to the outputs of the LSTM, TCN or Transformer architectures, ensuring a standardized decoding layer across different model families.

To maintain robustness against transient sensor noise or occasional misclassifications, the FSM ignores short, inconsistent phase fragments and segments labeled as auxiliary tasks. Its reliability is further enhanced by an internal step scoring mechanism that accumulates value based on the observation of a valid biomechanical sequence. In an ideal scenario, a complete sequence encompassing \textit{Take-off} $\to$ \textit{Swing} $\to$ \textit{Strike} $\to$ \textit{Stance} yields a maximal raw score of $4.0$, while partial but plausible sequences receive proportionally lower scores. This value is subsequently normalized to a range of $[0, 1]$. A step event is only officially emitted when this normalized score exceeds a predefined threshold $\alpha$, ensuring that the system requires sufficient cumulative evidence before declaring a step. When this threshold is reached, the FSM defines a discrete step interval, 
spanning from the first contributing phase to the final phase. This methodology makes the detection robust to isolated missing phases while preserving high temporal accuracy at the level of the overall step. Furthermore, the same FSM logic is utilized offline to process ground-truth labels, allowing for a direct and rigorous comparison between predicted step timelines and actual user intent during evaluation.

\section{Experiments}

\begin{table*}[]
\centering
\caption{\small Classification performance over three models}
\label{tb:performance}
\begin{tabular}{l cccccc ccc cc}\toprule
\multicolumn{1}{c}{\multirow{3}{*}{Model}} & \multicolumn{6}{c}{Phase detection - success rate (\%)}  & \multicolumn{3}{c}{\multirow{2}{*}{\begin{tabular}[c]{@{}c@{}}Step detection - \\ Success rate (\%)\end{tabular}}} & \multicolumn{2}{c}{\multirow{2}{*}{\begin{tabular}[c]{@{}c@{}}Runtime \\ (ms)\end{tabular}}} \\\cmidrule(lr){2-7}
\multicolumn{1}{c}{} & \multicolumn{3}{c}{w/o FSM} & \multicolumn{3}{c}{w/ FSM} &  & \\\cmidrule(lr){2-4}\cmidrule(lr){5-7}\cmidrule(lr){8-10}\cmidrule(lr){11-12}
\multicolumn{1}{c}{} & TS1 & TS2 & Total & TS1 & TS2 & Total & TS1 & TS2 & Total & PC & ES \\
\midrule
TCN              & 85 & 61 & 82 & 95 & 91 & \cellcolor[HTML]{C0C0C0}93 & 96 & 91 & \cellcolor[HTML]{C0C0C0}95 & \cellcolor[HTML]{C0C0C0} 0.98 $\pm$ 0.253 & \cellcolor[HTML]{C0C0C0} 1.9 $\pm$ 0.612 \\
LSTM             & 80 & 74 & 81 & 88 & 84 & 87 & 90 & 85 & 89 & 1.3 $\pm$ 0.278 & 11.7 $\pm$ 1.52 \\
Transformer      & 67 & 74 & 69 & 94 & 90 & \cellcolor[HTML]{C0C0C0}93 & 97 & 92 & \cellcolor[HTML]{C0C0C0}95 & 1.1 $\pm$ 0.366 & 4.3 $\pm$ 1.01 \\
\bottomrule
\end{tabular}
\vspace{-0.5cm}
\end{table*}

In this section, we evaluate the proposed approach for crutch-based gait detection and compare between the three models. The data collection and experiments were conducted with the approval of the ethics committee at Tel-Aviv University under application No. 0011588-2. 
Hyperparameter tuning was performed using Optuna \cite{Ozaki2025optunahub}, optimizing the model parameters across multiple configurations. 


\subsection{Dataset}

The training dataset was compiled from four healthy adult subjects who were instructed to perform crutch-assisted walking using their voluntarily chosen technique without specific gait instructions. Each subject completed an average of four laps on a 15 m straight path, including turning motions, with each lap lasting approximately 40 seconds. Following the hyper-parameter optimization of the pre-processing temporal windows, distinct configurations were established for each model architecture to maximize classification performance: the TCN utilized a window size of $h=8$ with a stride of $2$, resulting in $N=34,560$ labeled sequence samples; the LSTM employed a window size of $h=7$ with a stride of $5$, yielding $N=13,830$ samples; and the Transformer utilized a window size of $h=6$ with a stride of $1$, producing a dataset of $N=69,140$ labeled sequence samples. To evaluate the models' generalizability across different user profiles, a separate test set was collected from two distinct subjects: Test Subject 1 (TS1), a healthy able-bodied individual who completed six laps, and Test Subject 2 (TS2), an experienced exoskeleton subject with complete lower-limb paralysis who is functionally dependent on the ReWalk device for mobility and completed four laps. 


\subsection{Models Architectures}

To identify the architecture providing the optimal balance between classification accuracy and minimal inference latency, we compared between TCN, LSTM and a Transformer. The hyperparameter optimization for all models was performed to identify the configurations that minimize categorical cross-entropy loss. The TCN configuration utilizes two residual blocks with 96 feature channels, a kernel size of 2, and a spatial dropout of 0.255, optimized with a learning rate of $8.9 \times 10^{-4}$, totaling 67,398 trainable parameters. In contrast, the LSTM architecture consists of three bidirectional layers with 64 hidden units per direction, a dropout rate of 0.313, and a learning rate of $2.5 \times 10^{-4}$, totaling 233,734 trainable parameters. For the Transformer model, which contains 201,350 trainable parameters, the embedding maps into a $128$-dimensional space followed by two encoder layers, each with two attention heads and a 128-unit feed-forward network.

To map the extracted features to gait phase probabilities, all models incorporate a series of dense layers followed by a Softmax activation layer. Specifically, the TCN includes a fully connected layer with 96 units, while the Transformer utilizes a 64-unit dense head. To refine these predictions, an FSM is integrated at the output of all models to enforce biomechanically valid phase sequences. The FSM utilizes an internal normalized score to evaluate the progression of a step; a step is only officially registered if this score exceeds a predefined threshold of $\alpha = 0.6$, ensuring robustness against transient misclassifications or isolated sensor noise.


\subsection{Models Evaluation}

\subsubsection{Model Accuracy}

\begin{figure}
    \centering
    \includegraphics[width = 0.75\linewidth]{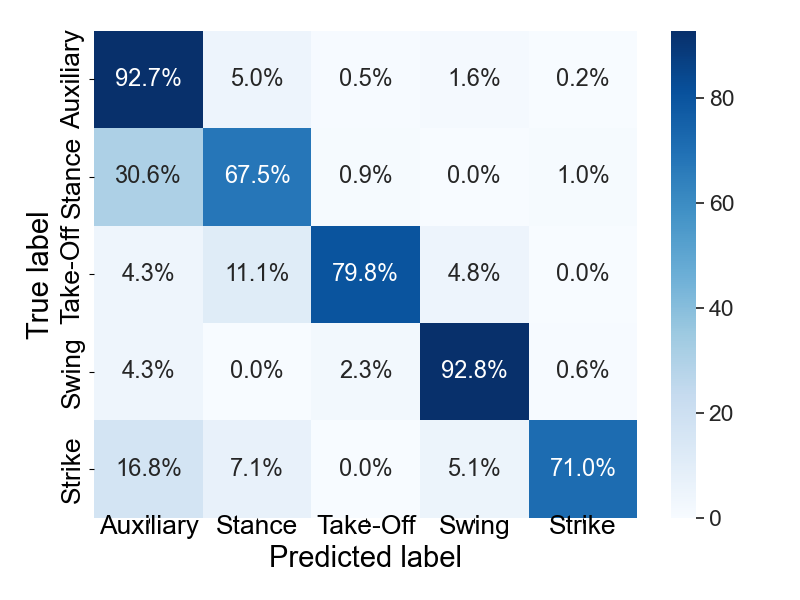} \\
    \vspace{-0.4cm}
    \caption{\small Confusion matrix for detecting the five crutch gait phases using the TCN and without using the FSM.}
    \label{fig:CM5}
    \vspace{-0.5cm}
\end{figure}

We first provide a performance comparison between the three classification models with and without including the FSM. Table \ref{tb:performance} presents the success rate results for phase and step detection over the test data. The results show individual success rates for TS1 and TS2, along with the total rates. First, we observe the phase detection accuracy. The results show a clear benefit of using the FSM, where it is able to significantly improve the detection accuracy of the five classes even for an ill-trained classification model. This is highlighted in the results for all three models, where they provide a rather poor predictions while the FSM significantly leveraged the detection accuracy. This performance discrepancy is elucidated by the confusion matrix in Figure \ref{fig:CM5}. The data reveals a clear classification bias where the models frequently misidentify the auxiliary phase as near-static locomotor phases, such as strike or stance. This suggests that the kinematic signatures of a stationary user performing auxiliary tasks share significant similarities with the grounded phases of active walking. To mitigate this overlap, future work should prioritize the inclusion of a more diverse and extensive auxiliary dataset to better refine the pattern boundaries between stationary activities and active locomotion. Nevertheless, the FSM is able to filter-out the non-locomotor phase and focus on gait phases. All three models demonstrate high accuracy in identifying the five gait phases, with the TCN and Transformer exhibiting a slight advantage in overall success rate. 

\begin{figure}
    \centering
    \includegraphics[width=1\linewidth]{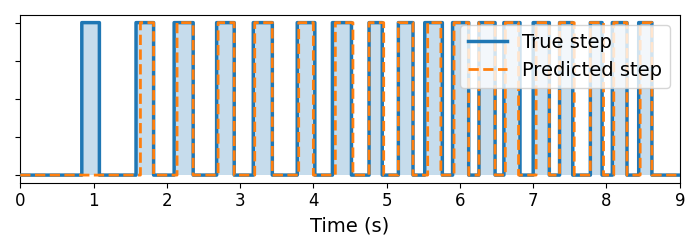}
    \vspace{-0.8cm}
    \caption{\small Real versus predicted steps of TS2 (exoskeleton user) based on observing crutch gait cycles.}
    \label{fig:steps}
    \vspace{-0.7cm}
\end{figure}

A primary application of the proposed approach is the synchronization of an assistive walking device's control cycle with user intent by monitoring crutch kinematics. Table \ref{tb:performance} also shows the success rates for detecting entire steps using the FSM. The results demonstrate that the system achieves strong temporal alignment, accurately capturing step occurrence and the specific timing of initiation and termination. A representative segment of this session is visualized in Figure \ref{fig:steps}. An analysis of missed detections reveals that the majority of classification errors occurred during gait initiation, specifically the first step taken from a static standing position. While the training set included labeled instances of these transitions, the model consistently struggled to recognize them compared to steady state walking cycles. This discrepancy suggests that the kinematic characteristics of the transition from static stance to locomotion differ significantly from those of continuous gait. Furthermore, the training data appears to lack sufficient volume of these first step samples to achieve high fidelity recognition. Consequently, a potential refinement for future iterations should involve adding initiation samples.

While training datasets were composed exclusively of data from healthy subjects, the performance for TP2 underscores the model high degree of cross domain generalizability. The architecture successfully identified the gait phases of an exoskeleton user with complete lower limb paralysis, demonstrating the potential for zero shot transfer to clinical populations. This can eliminate the need for intensive, patient specific retraining. Nevertheless, while current zero shot transfer is effective, including training data from paralyzed populations in future iterations is expected to further refine decision boundaries and maximize classification accuracy.

\subsubsection{Computational Efficiency}

Table \ref{tb:performance} details the computational runtime benchmarks for the models across two environments: a high-performance PC and an Embedded System (ES) representative of the target hardware for the walking device. The PC setup featured an Intel Core Ultra 7 (2.20 GHz) CPU with 32.0 GB of RAM. To evaluate the practical feasibility of real-time deployment on a walking device, runtimes were also measured on an ES, specifically the NVIDIA Jetson AGX Orin. To establish a baseline for processor-bound performance and simulate power-constrained operating conditions, inference was executed exclusively on the CPU in both instances, with no GPU acceleration utilized. The results highlight the TCN as the most advantageous architecture, as it simultaneously achieves the highest classification fidelity and the lowest inference latency across all tested platforms. Hence, we use the TCN for further analysis.

\subsubsection{Data Efficiency}

\begin{figure}
    \centering
    \includegraphics[width=1\linewidth]{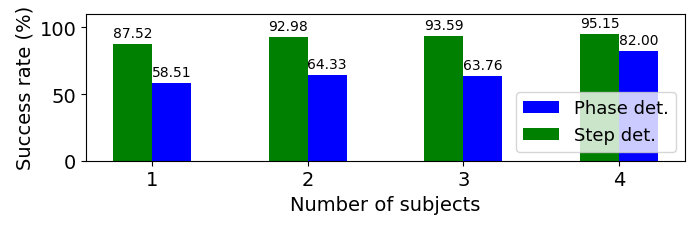}
    \vspace{-0.7cm}
    \caption{\small Classification success rate with regard to the number of subjects included in the training data, evaluated on the test data with the TCN model.}
    \label{fig:subjects}
    \vspace{-0.7cm}
\end{figure}

To determine the data requirements for the TCN model, we conducted an analysis on the size of the training population. This was performed through a cross-validation procedure where the TCN was iteratively trained using data from a varying number of subjects. For each size, the model was retrained across four unique subject combinations to compute an average success rate of phase and step detection, ensuring the results were not biased by a specific individual's gait characteristics. All resulting models were evaluated using the same test data from the two test subjects. As illustrated in Figure \ref{fig:subjects}, the average detection accuracy exhibits a positive correlation with the number of training subjects. However, a significant finding is the impact of the FSM on data efficiency. While the raw model accuracy improves with additional data, the integration of the FSM markedly lowers the data threshold required for effective operation; even with a single training subject, the FSM-augmented model achieves sufficient performance for practical use. Despite this baseline robustness, the results indicate that classification accuracy continues to scale with the inclusion of more diverse data, suggesting that further performance gains can be acquired by training with more subjects.

\section{Conclusions}

This research validates an approach for gait phase and step detection in lower limb walking devices using a single crutch mounted IMU. By identifying a five phase gait cycle without force sensing hardware, we provide a scalable, cost effective method for human exoskeleton coordination. The TCN emerged as the optimal architecture for embedded deployment, offering superior accuracy and minimal computational runtime. A core finding is the efficacy of the FSM as an augmentation layer, which significantly enhances accuracy and lowers the data threshold, achieving sufficient performance even with a single training subject. Furthermore, the model demonstrated high generalizability by accurately classifying the gait of a paralyzed user despite being trained solely on healthy subjects. This suggests that crutch kinematics provide a universal signature for intent recognition, facilitating clinical deployment without intensive patient specific retraining.

Future work should focus on enhancing gait initiation detection by introducing specialized phase labels to capture the unique kinematic transitions from standing to locomotion. In addition, one can aim to implement a high-level control loop that synchronizes leg gait phases with crutch motion, ensuring that mechanical assistance is perfectly phased with user intent. To facilitate this, future work may pursue full real-time embedded integration into a mobile walking device, optimizing the TCN architecture for power-constrained environments. Finally, systematic clinical trials are necessary to quantify objective improvements in gait stability, user experience, and safety compared to baseline exoskeleton operation.

\bibliographystyle{IEEEtran}
\bibliography{ref}

\end{document}